\documentclass[sigconf,screen,prologue,table,dvipsnames]{acmart}

\usepackage{booktabs} % For formal tables

\setcopyright{rightsretained}
\acmBooktitle{Indian Conference on Computer Vision, Graphics and Image
Processing (ICVGIP '21), December 19--22, 2021, Jodhpur, India}
\acmISBN{978-1-4503-7596-2}
\acmConference[ICVGIP'21]{12th Indian Conference on Computer Vision, Graphics and Image Processing}{December 2021}{Jodhpur, India}
\acmYear{2021}
\copyrightyear{2021}

\acmPrice{15.00}

\editor{Chetan Arora}
\editor{Parag Chaudhuri}
\editor{Subhransu Maji}

\usepackage{graphicx}
\usepackage{tabularx}
\usepackage{amssymb}
\usepackage{multirow}
\usepackage{siunitx}
\usepackage{xcolor,colortbl}
\usepackage{svg}

\definecolor{Gray}{gray}{0.85}
\definecolor{LightCyan}{rgb}{0.88,1,1}
\definecolor{White}{rgb}{1,1,1}
\definecolor{OliveGreen}{rgb}{0,0.6,0}
\definecolor{SoftRed}{rgb}{1,0.2,0.2}
\usepackage{hyperref}
\hypersetup{
     breaklinks=true,
     bookmarks=false,
     pagebackref=true,
     colorlinks=true,
     citecolor=OliveGreen,
     linkcolor= SoftRed,
     urlcolor= LightCyan
}
\usepackage{url}

\newcolumntype{g}{>{\columncolor{Gray}}c}
\makeatletter
\newcommand{\printfnsymbol}[1]{%
  \textsuperscript{\@fnsymbol{#1}}%
}

\acmDOI{10.1145/3490035.3490270}

\acmArticle{13}

\begin{document}

% UPDATE the title, author names, affiliations and shortauthors 
\title{NTU-X: An Enhanced Large-scale Dataset for Improving Pose-based Recognition of Subtle Human Actions}
%\titlenote{Produces the %permission block, and
%  copyright information}

  \author{Neel Trivedi}
  %\authornote{Dr.~Trovato insisted his name be first.}
%   \orcid{1234-5678-9012}
  \affiliation{%
    \institution{CVIT, IIIT-Hyderabad}
    % \streetaddress{P.O. Box 1212}
    \city{Hyderabad 500032}
    \country{INDIA}
  }
  \email{neel.trivedi@research.iiit.ac.in}
  
   \author{Anirudh Thatipelli}
  \authornote{Equal contribution}
%   \orcid{1234-5678-9012}
  \affiliation{%
    \institution{CVIT, IIIT-Hyderabad}
    % \streetaddress{P.O. Box 1212}
    \city{Hyderabad 500032}
    \country{INDIA}
  }
  \email{at794@snu.edu.in}
  
   \author{Ravi Kiran Sarvadevabhatla}
  %\authornote{Dr.~Trovato insisted his name be first.}
%   \orcid{1234-5678-9012}
  \affiliation{%
    \institution{CVIT, IIIT-Hyderabad}
    % \streetaddress{P.O. Box 1212}
    \city{Hyderabad 500032}
    \country{INDIA}
  }
  \email{ravi.kiran@iiit.ac.in}
  
%   \author{Second Author}
%   %\authornote{The secretary disavows any knowledge of this author's actions.}
%   \affiliation{%
%     \institution{Institute for Clarity in Documentation}
%     \streetaddress{P.O. Box 1212}
%     \city{Dublin}
%     \state{Ohio}
%     \country{USA}
%     \postcode{43017-6221}
%   }
%   \email{webmaster@marysville-ohio.com}

\renewcommand{\shortauthors}{}

\begin{abstract}
The lack of fine-grained joints (facial joints, hand fingers) is a fundamental performance bottleneck for state of the art skeleton action recognition models. Despite this bottleneck, community's efforts seem to be invested only in coming up with novel architectures. To specifically address this bottleneck, we introduce two new pose based human action datasets - NTU60-X and NTU120-X. Our datasets extend the largest existing action recognition dataset, NTU-RGBD. In addition to the 25 body joints for each skeleton as in NTU-RGBD, NTU60-X and NTU120-X dataset includes finger and facial joints, enabling a richer skeleton representation. We appropriately modify the state of the art approaches to enable training using the introduced datasets. Our results demonstrate the effectiveness of these NTU-X datasets in overcoming the aforementioned bottleneck and improve state of the art performance, overall and on previously worst performing action categories. Code and pretrained
models can be found at \url{https://github.com/skelemoa/ntu-x}.
\end{abstract}

% UPDATE THIS



\maketitle

\section{Introduction}
\label{sec:intro}

\begin{figure}[]
    \centering
    \includegraphics[width=\linewidth]{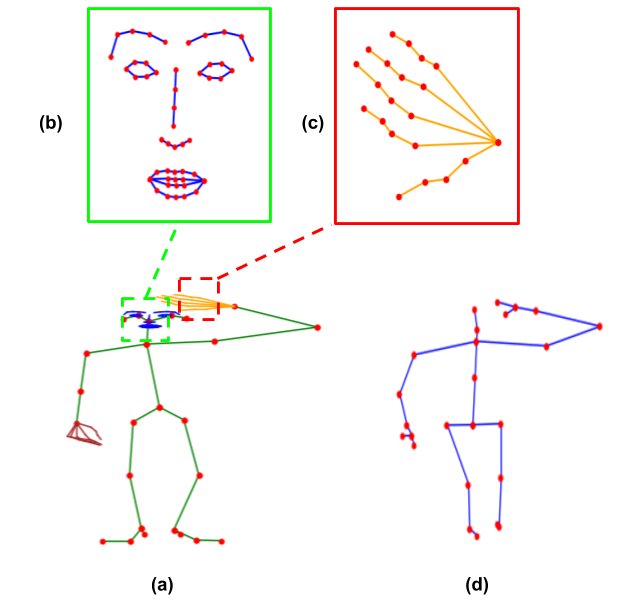}
    \caption{(a) The $118$ joint skeleton introduced in the new NTU-X datasets. The $25$ body joints are indicated by red dots.(b) $51$ facial joints (c) $21$ finger joints (d) $25$ body joints present in original NTU datasets .  }
    \label{fig:joints_fig}
\end{figure}

Understanding human actions from visual data is crucial for applications related to surveillance, interactive user interfaces and multimedia systems. This task is usually accomplished with RGB videos as input~\cite{Poppe_IV_2010}. However, advances in technologies have enabled use of other modalities (e.g. depth~\cite{wang2012mining}) and systems such as Microsoft Kinect which can provide skeleton-like human pose representations~\cite{ren2020survey}. In particular, the introduction of large-scale skeleton-based human action datasets NTU RGB+D~\cite{Shahroudy_2016_CVPR,Liu_2019_NTURGBD120} has shifted the focus towards skeleton-based human action recognition approaches. In contrast to full-frame RGB-based representations, 3D skeleton joints encode human body dynamics in a computationally efficient manner, preserve privacy and can offer greater robustness to view and illumination.

\begin{figure*}[!t]
    \centering
    \includegraphics[width=\textwidth]{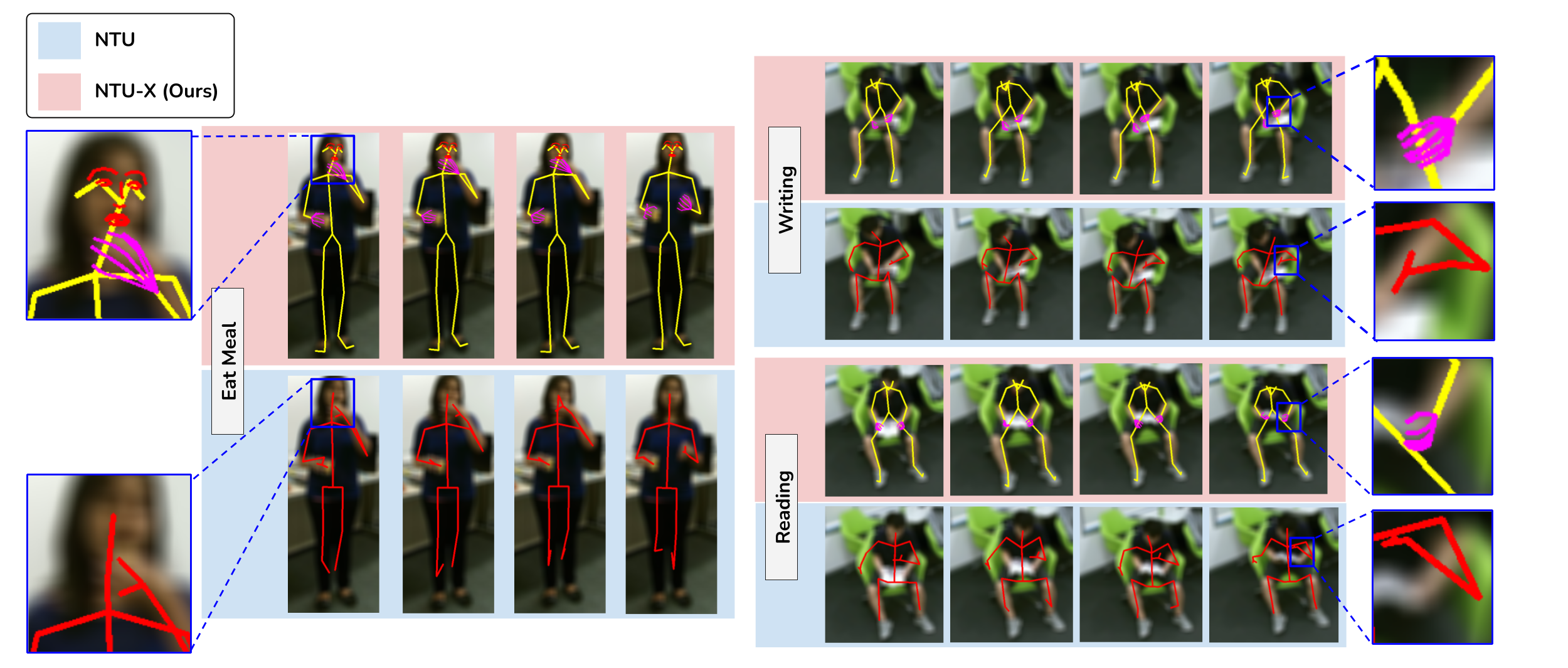}
    \caption{Sample skeletons from original NTU Kinect dataset (blue background) and proposed NTU-X dataset (pink background). Note that blurred RGB frame is included only for reference and is not part of skeleton data. The three classes mentioned - `eat meal', `writing' and `reading' are few of the most confused classes for NTU dataset (see Table~\ref{tab:bottom5ntu60}). As the zoomed insets illustrate, the quality of joints captured by NTU-X dataset is better compared to the original NTU dataset.}
    \label{fig:seq_fig}
\end{figure*}

The recent adoption of graph neural networks which process the skeleton action sequence as a  spatio-temporal graph has enabled a steady rise in average accuracy for skeleton action recognition~\cite{stgcn2018aaai,liu2020disentangling,cheng2020shiftgcn,song2020stronger,dstanet_accv2020}. However, an analysis of sorted per-class accuracies reveals that the actions with lowest accuracy involve the usage of fingers (see Table~\ref{tab:bottom5ntu60},\ref{tab:bottom5ntu120}). The underlying reason is that hand joints in Kinect-based skeletons provided in the original dataset are represented by just two finger joints (Figure~\ref{fig:joints_fig}-d). As a result, actions involving subtle finger movements (e.g. `eating', `writing', 'make ok sign', 'make victory sign') often fail to be recognized correctly. Sometimes, even the non-hand, main body joints are localized poorly by the Kinect-based capture system, as Figure~\ref{fig:seq_fig} shows. These shortcomings at raw data level cannot be addressed at the architecture level, i.e. by proposing novel architectures.

To address the mentioned data-level issues, we introduce NTU60-X and NTU120-X, curated and extended versions of the existing NTU  dataset. Obtained from RGB videos present along with NTU skeleton data, the pose representations in the new dataset include $42$ finger joints ($21$ for each hand), $51$ facial keypoint joints and $25$ body joints similar to those present in Kinect-based NTU-60 and NTU-120, for a total of $118$ joints per skeleton (see Figure~\ref{fig:joints_fig}). We also modify state of the art approaches to enable experimental evaluation and benchmark the modified variants on NTU60-X and NTU120-X. As a result, we set the new state of the art benchmark on NTU-60 and NTU-120. Our results also demonstrate the benefit of the newly introduced datasets for overcoming the performance bottleneck mentioned earlier and enabling recognition of subtle human actions involving hand-based joints. 
Resources (source code, pre-trained models, analysis and videos) related to NTU-X are available at  \url{https://github.com/skelemoa/ntu-x}.

\section{Related Work}
\label{sec:related}

Prior to creation of NTU RGB-D dataset, a number of datasets enabled progress for skeleton-based human action recognition. MSR-Action3d~\cite{li2010action} was one of the first action recognition datasets which provided depth and  skeleton joint modalities, albeit from a single viewpoint. However, it only covered a limited set of gaming actions (e.g. forward punching, side boxing). The Northwestern-UCLA dataset~\cite{wang2014crossview} scaled up the diversity to include videos from multiple views and with actions performed by $10$ different actors. The NTU RGB-D 60 dataset~\cite{Shahroudy_2016_CVPR} comprises of $60$ action categories, performed by $40$ subjects. Its extension NTURGB-D 120 ~\cite{Liu_2019_NTURGBD120} is one of the largest and most diverse skeleton action dataset comprising of $120$ actions performed by $106$ subjects from $155$ viewpoints.

 Varying-view RGB-D Action Dataset (VAD)~\cite{ji2019large} comprises view-varying Kinect captured sequences covering the entire \ang{360} view angles, containing $40$ actions that are performed by $118$ distinct performers. Unfortunately, the full dataset is not publicly available (as of current). Notably, the datasets mentioned above do not provide fine-grained joints for hands and faces which limits their utility for certain actions as mentioned previously. 
 
 An alternative approach for skeleton estimation infers the joints from RGB video frames without requiring specialized capture equipment. In the Kinetics-skeleton dataset~\cite{stgcn2018aaai}, the 2D skeleton joint coordinates predicted from RGB frames are combined with the joint estimation confidence to obtain a pseudo 3D skeleton representation on videos from Kinetics-400 action dataset~\cite{Carreira2017QuoVA}. However, the resulting skeleton dataset contains many invalid sequences~\cite{Gupta2021}.  
 
 We summarize the salient aspects of these datasets and our proposed NTU60-X and NTU120-X in Table~\ref{tab:dataset}.

\begin{table*}[!t]
\resizebox{0.9\textwidth}{!}
 {%
  \centering 
 \begin{tabular}{l|c|c|c|c|c|c}
 \toprule
             Dataset & Body & Face & Fingers & Sequences & Classes & Joints \\    
 \midrule
   MSR-Action3D\cite{li2010action} & \checkmark & & & $567$ & $20$ & $20$\\
   Northwestern-UCLA\cite{wang2014crossview} & \checkmark & & & $1{,}475$ & $10$ & $21$\\
  VAD\cite{ji2019large} & \checkmark & & & $25{,}600$ & $40$ & $25$ \\
  NTU RGB+D\cite{Shahroudy_2016_CVPR} & \checkmark & & & $56{,}880$ & $60$ & $25$ \\
  NTU RGB+D 120\cite{Liu_2019_NTURGBD120} & \checkmark & & & $114{,}035$ & $120$ & $25$ \\
  \rowcolor{Gray}
  \textbf{NTU60-X (Ours)} & \checkmark & \checkmark & \checkmark & $56{,}148$ & $60$ & $118$ \\
  \rowcolor{Gray}
  \textbf{NTU120-X (Ours)} & \checkmark & \checkmark & \checkmark & $113{,}821$ & $120$ & $118$ \\
  \bottomrule
 \end{tabular}
  }
  
\caption{ Comparison between NTU-X and some of the other publicly available skeleton-action recognition datasets. We are one of the first datasets to include body, face and hands joints in 3D for multi-person and occlusion case as well.}
\label{tab:dataset}
\end{table*}
%%%%%%%%%%%%%%%%%%%%%%%%%%%%%%%%%%%%%%%%%%%%%%%%%%%%%%%%%%%%%%%%%%%%%%%%%%%%%%%%

\section{NTU-X}
\label{sec:ntu60x}

\begin{figure*}[]
    \centering
\includegraphics[width=0.9\textwidth]{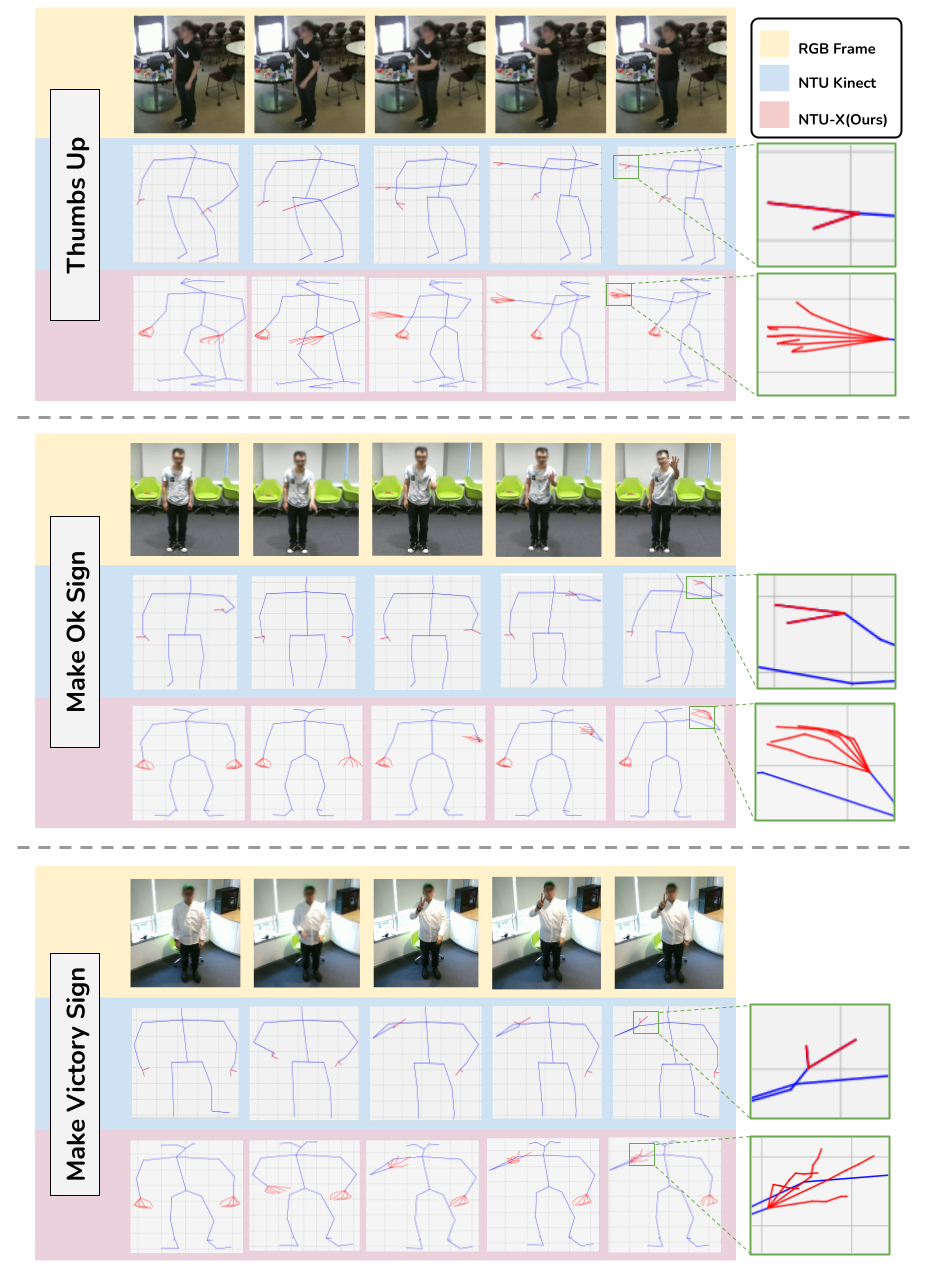}
    \caption{Samples showing 3d plot of the original NTU kinect skeletons and newly proposed NTU-X skeletons with corresponding RGB frames. The zoomed insets show the finger joints estimated in both NTU-Kinect and NTU-X and it clearly shows that NTU-X represents the action much more comprehensively than original NTU-Kinect data.}
    \label{fig:mega_fig}
\end{figure*}

The NTU RGB+D dataset~\cite{Shahroudy_2016_CVPR} provides RGB videos along with 3D Kinect skeleton data. We first extract the RGB frames from the videos at the frame rate of 30 FPS. We estimate 3D poses from RGB frames using SMPL-X~\cite{SMPL-X:2019}. SMPL-X uses strong 2D pose priors estimated using Openpose~\cite{cao2018openpose} on each RGB frame. However, SMPL-X based pose estimation is rather slow and is reliant on optimization heuristics. It also fails on blurred images and in the presence of light occlusion. To compensate for these issues, we use ExPose~\cite{ExPose:2020}. ExPose uses a part-wise attention-based model that feeds high resolution patches of the corresponding body parts to their dedicated refinement module. Unlike SMPL-X, ExPose estimates the full 3D pose (body, finger and face joints) from the RGB image without relying on 2D pose prior and is much faster compared to SMPL-X.

Since it is difficult to automatically select between SMPL-X and ExPose pose representations, we employ a semi-automatic approach to curate the final dataset. We use Openpose~\cite{cao2018openpose} toolbox to estimate the 2D pose and associated confidence for the full-body joints. Openpose provides total $70$ joints for face out of which we use $51$ major joints as shown in Figure~\ref{fig:joints_fig}(b) to make the final skeleton of $118$ joints ($25$ body + $21 \times 2$ fingers + $51$ face) for each frame of the clips. Keeping the intra-view and intra-subject variance of the NTU dataset in mind, we sample random videos covering each view per class of NTU and estimate the SMPL-X and ExPose outputs. We examine the quality of the skeleton backprojected to RGB frame and use the accuracy of alignment to select between ExPose and SMPL-X. Empirically, we observe that ExPose and SMPL-X perform equally well for single-person actions but SMPL-X, though slow, provides better pose estimates for multi-person action class sequences. To check which pose extraction method (ExPose or SMPLx) has been used for each of the classes in the NTU-X dataset, kindly refer to our Github project at \url{https://github.com/skelemoa/ntu-x}.
%For the multi-person case, we select the two people with the highest average joint confidence estimated using OpenPose for each clip. Subsequently, we obtain a 118 joints 3D skeleton (X, Y, Z) for each person per frame. Subsequently, we represent each clip of T frames with tensors of (3, T, 118, 2) dimensions. Keeping T as 300, we pad each clip with the remaining frames as zeroes.

To ensure good dataset quality, we remove corrupted videos from the original dataset, using a procedure similar to one adopted for the original dataset~\cite{Shahroudy_2016_CVPR}. We also omit videos in which people are completely absent. Additionally, for some  samples  OpenPose provides poor estimates and hence we discard instances of such videos as well.

%%%%%%%%%%%%%%%%%%%%%%%%%%%%%%%%%%%%%%%%%%%%%%%%%%%%%%%%%%%%%%%%%%%%%%%%%%%%%%%%

\section{Experiments}
\label{sec:experiments}

To evaluate the impact of NTU60-X and NTU120-X on overall performance, we benchmarked models with state-of-the-art performance on NTU60 and NTU120. We selected DSTA-Net\cite{dstanet_accv2020}, 4s-ShiftGCN~\cite{cheng2020shiftgcn}, MS-G3D~\cite{liu2020disentangling} and PA-ResGCN~\cite{song2020stronger} as the models to benchmark the newly introduced datasets. For the models DSTA-Net\cite{dstanet_accv2020}, MS-G3D~\cite{liu2020disentangling} and 4s-ShiftGCN~\cite{cheng2020shiftgcn}, we updated the graph structure of the skeletons to incorporate the newly introduced joints. Figure~\ref{fig:joints_fig}(d) shows the skeleton topology for the original kinect data. We changed the input graph topology for these models according to our new skeleton structure as shown in Figure~\ref{fig:joints_fig}(a).

PA-ResGCN~\cite{song2020stronger}, being a semantic part-based model, required more significant modification. Along with changes in input skeleton graph structure as done for the other two models, we defined new parts to incorporate the newly introduced joints and thus enable richer feature extraction. Since this model learns attentive weights for each of the input skeleton joints by dividing the skeleton into different parts, the definition of parts were also changed based on the NTU-X skeleton. In case of NTU-RGBD skeleton, PA-ResGCN defines total $5$ parts: torso, left arm, right arm, left leg and right leg. In the new NTU-X skeleton, $3$ additional parts were defined for 67 joints (body + fingers) skeleton: left fingers, right fingers and head, resulting in a total of $8$ parts. For 118 joint (body + fingers + face) skeleton along with these $3$ additional parts, one more part of face was added resulting in a total of $9$ parts. 

\subsection{Results}
\label{subsec:results}

\begin{table*}[]
\resizebox{0.8\textwidth}{!}
 {%
  \centering 
 \begin{tabular}{l|cgcg}
  \toprule
   model & NTU60 & NTU60-X (Ours) & NTU120 & NTU120-X (Ours) \\
   \midrule
   DSTA-Net\cite{dstanet_accv2020} & 91.50 &     \cellcolor{LightCyan}\textbf{93.56} & 86.60 & \cellcolor{LightCyan}\textbf{87.80}\\
    4s-ShiftGCN\cite{cheng2020shiftgcn} & 90.70 & \textbf{91.78} & 85.90 & \textbf{86.18}\\
   MS-G3D\cite{liu2020disentangling} & 91.50 & \textbf{91.76} & 86.9 & \textbf{87.10}\\
   PA-ResGCN\cite{song2020stronger} & 90.9 & \textbf{91.64} & \textbf{87.4} & 86.42\\
  \bottomrule
 \end{tabular}
  }
\caption{\label{tab:results_ntux} Results for top performing models of NTU60 and NTU120 dataset on NTU60-X dataset and NTU120-X (with finger joints) - see Section~\ref{subsec:results}. The gray shaded columns show results on our newly introduced dataset. The blue highlighted cell corresponds to  best overall performance for 60 and 120 class setups.}
\end{table*}

The results of training the four selected models on the new NTU60-X and NTU120-X datasets, with finger joints included for Cross Subject protocol, are shown in Table ~\ref{tab:results_ntux}. Clearly, our modified DSTA-Net, MS-G3D, 4s-Shift-GCN and PA-ResGCN outperform their counterparts' performance on the original NTU60 dataset by a significant margin. For NTU120-X, all three models except PA-ResGCN outperform their counterparts' performance on NTU120 dataset. PA-ResGCN fails to surpass the orginal accuracy for 120 class dataset by a small margin. We hypothesize that this could be due to PA-ResGCN's architecture being too specific for the original Kinect skeleton setup and unable to handle the addition of extra added finger joints in the large-category (120 class) setting. We can also see that DSTA-Net not only beats its numbers on the original dataset, but also achieves state of the art performance among all the models with a margin of more than 2\% for NTU60 dataset.(See highlighted cells in Table~\ref{tab:results_ntux}). 

These results also support the fact that existing approaches, if provided better and richer joint data, have the capacity to perform better. A detailed analysis of each model's performance and category level improvements is discussed next.

\subsection{Discussion}
\label{subsec:discussion}

\begin{table}[]
\resizebox{\linewidth}{!}
 {%
  \centering 
 \begin{tabular}{c|c|cg}
 \toprule
            \rowcolor{White} 
            Model & Class name & NTU-60 & NTU60-X\\    
 \midrule
    
    \multirow{5}{4em}{\small \centering \textbf{DSTA-Net \cite{dstanet_accv2020}}} & Writing & 67.04 \% & 79.41 \% \\
    & Reading & 68.75 \% & 94.49 \%  \\
    & Play with phone/tablet & 71.06 \% & 86.91 \% \\
    & Type on a keyboard & 71.64 \% & 93.45 \% \\
    & Sneeze or cough & 73.55 \% & 80.07 \% \\ 
    \midrule
    
    \multirow{5}{4em}{\small \centering \textbf{4s-ShiftGCN \cite{cheng2020shiftgcn}}} & Writing & 65.19 \% & 76.23 \%\\
    & Reading & 68.75 \% & 91.91 \%\\
    & Eat meal & 72.89 \% & 80.22 \%\\
    & Type on keyboard & 73.82 \% & 91.24 \% \\
    & Sneeze or cough & 73.91 \% & 75.72 \% \\
    \midrule
    
    \multirow{5}{4em}{\small \centering \textbf{MS-G3D \cite{liu2020disentangling}}} & Writing & 57.41 \% & 72.96 \%\\
    & Eat meal & 71.43 \% & 79.85 \%\\
    & Reading & 72.43 \% & 92.28 \%\\
    & Sneeze or cough & 77.17 \% & 80.80 \% \\
    & Play with phone/tablet & 78.75 \% & 79.41 \%\\
    
    \midrule\multirow{5}{4em}{\small \centering \textbf{PA-ResGCN \cite{song2020stronger}}} & Writing & 63.97 \% & 78.89 \%\\
    & Reading & 67.65 \% & 94.12 \%\\
    & Sneeze or cough & 73.91 \% & 76.45 \% \\
    & Type on keyboard & 74.91 \% & 91.61 \% \\
    & Eat meal & 74.91 \% & 80.95\%\\
  \bottomrule
 \end{tabular}
  }
\caption{The NTU60 column shows accuracies of bottom 5 action classes for models trained on original NTU60 dataset. The NTU60-X column shows accuracies of the same classes but with models trained on our NTU60-X dataset (finger joints: Section~\ref{sec:ntu60x}). Thanks to availability of additional finger joint information in NTU-60X, we see visible performance improvement across all the models.}
\label{tab:bottom5ntu60} 
\end{table}

\begin{table}[]
\resizebox{\linewidth}{!}
 {%
  \centering 
 \begin{tabular}{c|c|cg}
 \toprule
            \rowcolor{White} 
            Model & Class name & NTU-120 & NTU120-X\\    
 \midrule
    
    \multirow{5}{4em}{\small \centering \textbf{DSTA-Net \cite{dstanet_accv2020}}} & Staple book & 37.65 \% & 36.60 \% \\
    & Make ok sign & 51.13 \% & 72.70 \%  \\
    & Make victory sign & 53.85 \% & 60.49 \% \\
    & Counting money & 59.65 \% & 84.91 \% \\
    & Blow nose & 64.94 \% & 70.73 \% \\ 
    \midrule
    
    \multirow{5}{4em}{\small \centering \textbf{4s-ShiftGCN \cite{cheng2020shiftgcn}}} & Staple book & 35.9 \% & 34.19 \%\\
    & Make victory sign & 53.32 \% & 62.59 \%\\
    & Make ok sign & 55.83 \% & 64.35 \%\\
    & Counting money & 61.93 \% & 82.11 \% \\
    & Blow nose & 63.07 \% & 67.60 \% \\
    \midrule
    
    \multirow{5}{4em}{\small \centering \textbf{MS-G3D \cite{liu2020disentangling}}} & Staple book & 32.57 \% & 34.50 \%\\
    & Make victory sign & 54.02 \% & 68.18 \%\\
    & Hit with object & 60.03 \% & 69.98 \%\\
    & Blow nose & 60.45 \% & 68.64 \% \\
    & Counting money & 60.70 \% & 90.53 \%\\
    
    \midrule\multirow{5}{4em}{\small \centering \textbf{PA-ResGCN \cite{song2020stronger}}} & Staple book & 40.63 \% & 36.08 \%\\
    & Make victory sign & 59.79 \% & 60.49 \%\\
    & Hit with object & 61.90 \% & 63.53 \% \\
    & Cutting paper & 63.30 \% & 55.32 \% \\
    & Counting money & 64.65 \% & 78.25 \%\\
  \bottomrule
 \end{tabular}
  }
\caption{The NTU120 column shows accuracies of bottom 5 action classes for models trained on original NTU120 dataset. The NTU120-X column shows accuracies of the same classes but with models trained on our NTU120-X dataset (finger joints: Section~\ref{sec:ntu60x}). Thanks to availability of additional finger joint information in NTU-120X, we see visible performance improvement across all the models.}
\label{tab:bottom5ntu120} 
\end{table}

Table~\ref{tab:bottom5ntu60} and Table~\ref{tab:bottom5ntu120} list the five worst performing classes for all the four models on the original NTU60 and NTU120 datasets respectively along with their per class accuracy. The shaded columns in these tables provide the accuracy of these classes when the models are trained using the newly introduced NTU60-X and NTU120-X dataset. From these results, it is evident that most of the bottom performing classes for the original NTU datasets involve actions with fine finger movements (e.g. ``writing", ``type on keyboard", ``eat meal", ``make ok sign", ``make victory sign"). When the models are provided with input data that includes finger joints, the per class accuracy for such categories is improved significantly. Figure~\ref{fig:seq_fig} and Figure~\ref{fig:mega_fig} also show that without inclusion of finger level joints, recognition of such action categories is ambiguous and difficult.

\begin{table*}[]
\resizebox{0.9\linewidth}{!}
 {%
  \centering 
 \begin{tabular}{c|c|c}
 \toprule
             & \textbf{Bottom-5 NTU60-X} & \textbf{Bottom-5 NTU120-X}\\    
 \midrule
    \multirow{5}{5em}{\textbf{\centering DSTA-Net
    \cite{dstanet_accv2020}}} 
    & Writing (79.41 \%) & Staple book (36.6 \%)\\
    & Eat meal (80.0 \%) & Make victory sign (60.49 \%)\\
    & Sneeze or cough (80.07 \%) & Cutting paper (61.26 \%)\\
    & Touch head (headache) (80.73 \%) & Fold paper (68.17 \%)\\
    & Take off a shoe (84.67 \%) & Play magic cube (69.47 \%)\\
    \midrule
    \multirow{5}{5em}{\textbf{\centering 4s-ShiftGCN
    \cite{cheng2020shiftgcn}}}
    & Sneeze or cough (75.72 \%) & Staple book (34.15 \%)\\
    & Writing (76.3 \%) & Cutting paper (52.01 \%)\\
    & Eat meal (80.22 \%) & Playing with phone/tablet (60.73 \%)\\
    & Touch head (headache) (80.36 \%) & Fold paper (61.22 \%)\\
    & Playing with phone/tablet (80.51 \%) & Make victory sign (62.59 \%)\\
    \midrule
    \multirow{5}{5em}{\textbf{\centering MS-G3D
    \cite{liu2020disentangling}}}
    & Writing (72.59 \%) & Staple book (34.5 \%)\\
    & Eat meal (79.12 \%) & Fold paper (61.91 \%)\\
    & Wear a shoe (80.07 \%) & Cutting paper (63.18 \%)\\
    & Punching/slapping other person (80.66 \%) & Playing with phone/tablet (64.73 \%)\\
    & Sneeze or cough (81.52 \%) & Make ok sign (65.91 \%)\\
    \midrule
    \multirow{5}{5em}{\textbf{\centering PA-ResGCN
    \cite{song2020stronger}}}
    & Sneeze or cough (76.45 \%) & Staple book (38.53 \%)\\
    & Writing (78.89 \%) & Cutting paper (55.32 \%)\\
    & Touch head (headache) (78.91 \%) & Make victory sign (61.19 \%)\\
    & Punching/slapping other person (80.66 \%) & Playing with phone/tablet (61.45 \%)\\
    & Eat meal (80.95 \%) & Fold paper (62.78 \%)\\
  \bottomrule
 \end{tabular}
  }
\caption{\label{tab:bottom5ntux} This table shows the bottom performing classes for all the models evaluated in this paper on the newly introduced NTU60-X and NTU120-X datasets. This table clearly indicates that the overall accuracy of bottom performing classes for the newly introduced NTU60-X and NTU120-X are higher than the overall accuracy of bottom performing classes for the orginal datasets, as shown in Table~\ref{tab:bottom5ntu60} and Table~\ref{tab:bottom5ntu120}}
\end{table*}

Table~\ref{tab:bottom5ntux} shows the worst performing classes for the NTU-X dataset for all the models. Comparing with  accuracy of worst performing classes of the original NTU dataset given in Table~\ref{tab:bottom5ntu60} and Table~\ref{tab:bottom5ntu120}, we see that even the accuracy of worst performing classes of NTU-X dataset is, on average, higher compared to the original NTU dataset.

\begin{figure*}[]
    \centering
    \includegraphics[width=0.95\textwidth]{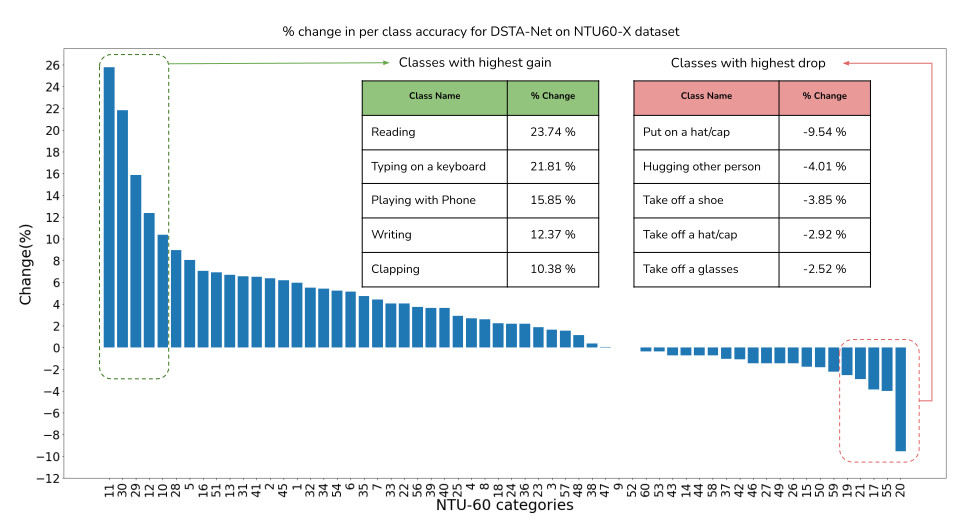}
    \caption{The \% gain in per class accuracy for best performing model (DSTA-Net) after training on newly introduced NTU60-X dataset. The x-axis shows category id. The inset tables show actions with largest and least gain.}
    \label{fig:ntu60_change}
\end{figure*}

\begin{figure*}[]
    \centering
    \includegraphics[width=0.95\textwidth]{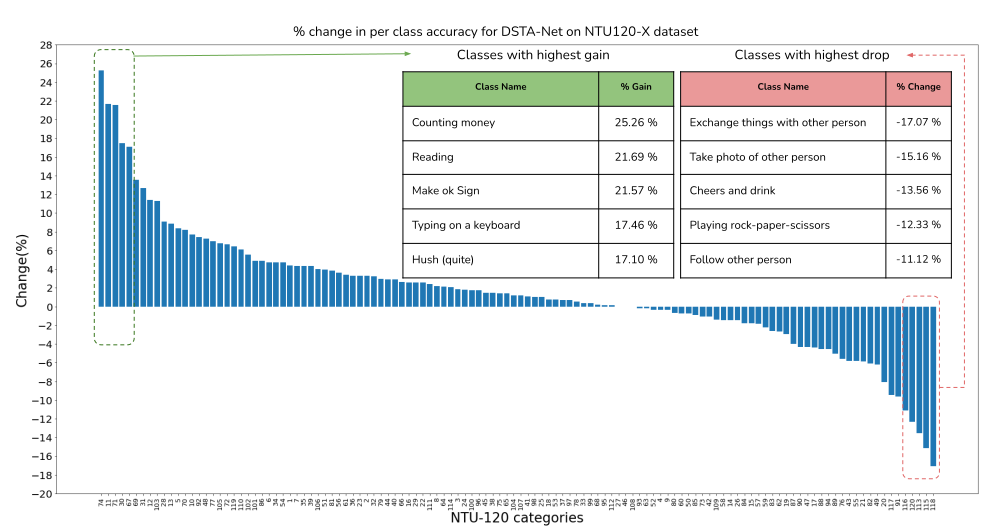}
    \caption{The \% gain in per class accuracy for best performing model (DSTA-Net) after training on newly introduced NTU120-X dataset. The x-axis shows category id. The inset tables show actions with largest and least gain.}
    \label{fig:ntu120_change}
\end{figure*}

To further illustrate the performance boost we gain by including the finger level joints into the input skelton, we show the change in per class accuracy when going from orginal NTU dataset to newly introduced NTU-X dataset in Figure~\ref{fig:ntu60_change} and Figure~\ref{fig:ntu120_change} for the top performing model DSTA-Net\cite{dstanet_accv2020}(based on Table~\ref{tab:results_ntux}). 

The table provided in the inset of Figure~\ref{fig:ntu60_change} shows classes which are benefited the most when going from NTU60 to NTU60-X dataset for the state-of-the-art performer (DSTA-Net). It is easy to see that these classes predominantly involve finger level actions such as ``reading", ``typing on keyboard", ``playing with phone" and ``writing". One can also observe that the gain for these classes is as high as \textbf{10-23 \%}. A similar trend can be seen in Figure~\ref{fig:ntu120_change} which shows classes which benefit the most when going from NTU120 to NTU120-X dataset. Once again, the classes with highest gain involve finger level actions (e.g. ``make of sign",``hush") and the gain for these classes is in the range \textbf{17-25\%}. Both of these tables clearly indicate training recognition approaches on our proposed NTU-X dataset significantly boosts the performance of classes involving finger movements.

From Figure~\ref{fig:ntu60_change} and Figure~\ref{fig:ntu120_change}, we also note that training on NTU-X dataset (finger joints) is not beneficial for all the classes, with a performance drop seen in some cases. Most of these classes involve another person (e.g. ``hugging other person", ``take photo of other person", ``playing rock-paper scissors" etc.). As per our understanding, capturing accurate skeletons for multiple people in a RGB frame is difficult which leads to poor pose extraction and ambiguity in classifying the sequences involving multiple people.

% or they are highly dependent on interaction with some object (e.g. ``put on hat", ``take off shoe", ``take off glasses" etc.). And since skeleton data doesn't contain any information regarding the object, these classes are suffering a drop in accuracy. 

However, it is clear that the overall magnitude of gain is higher than the magnitude of drop in per class accuracy. Hence, the average accuracy is higher for NTU-X dataset than the original NTU dataset.

\subsection{Ablation Study}
\label{sec:ablation}

\begin{table*}[h!]
\resizebox{0.9\textwidth}{!}
 {%
  \centering 
 \begin{tabular}{l|cgc|cg}
 \toprule
    & \multicolumn{3}{c|}{NTU60-X} & \multicolumn{2}{c}{NTU120-X}\\
   \midrule
   model & body & body + fingers & body + fingers + face & body & body + fingers\\
   \midrule
   DSTA-Net\cite{dstanet_accv2020}$^{*}$ & 89.69 & \textbf{90.91} & 88.97 & 84.82 & \textbf{87.80}\\
   4s-ShiftGCN\cite{cheng2020shiftgcn} & 89.56 & \textbf{91.78} & 89.64 & 84.78 & \textbf{86.18}\\
   MS-G3D\cite{liu2020disentangling} & 91.26 & \textbf{91.76} & 91.12 & 84.50 & \textbf{87.10}\\
   PA-ResGCN\cite{song2020stronger} & 89.98 & \textbf{91.64} & 89.79 & 82.85 & \textbf{86.42}\\
  \bottomrule
 \end{tabular}
  }
\caption{\label{tab:face-finger-comparison_ntu60x} Results on different variants of NTU60-X and NTU120-X dataset to understand the contribution of the additional joints. (*: Ablations on DSTA-Net are done using only the Joint stream of the network which contributes most to its performance.)}
\end{table*}

To examine the importance of body joints, finger joints and face joints individually, we also perform experiments with only body joints ($25$ joints), body + finger joints ($67$ joints) and body + fingers + face joints ($118$ joints) as well. The results of ablation study are shown in Table ~\ref{tab:face-finger-comparison_ntu60x}. The performance degrades when face joints are included with the body and finger joints. One reason for this could be that the actions in NTU dataset do not involve significant facial motion. Hence, the additional joints of the face make the skeleton graph larger than necessary and difficult for model optimization. Another possible reason could be that the existing models do not have a suitable architecture to handle the dense subgraph arising from the presence of facial keypoints. The poor results of models trained with only body joints ($25$ joints) are also in line with our hypothesis that the inclusion of finger joints in the input skeleton is crucial for better performance. In other words, the performance gain is not merely due to the shift from Kinect-based to RGB-based skeleton generation process.

\section{Conclusion}
\label{sec:conclusion}

In this paper, we have shown that the lack of hand-level joints is a fundamental performance bottleneck in the skeleton data of the largest action recognition dataset, NTU-RGBD. To address this bottleneck, we contribute a carefully curated skeleton dataset which provides finger-level hand joints and facial keypoint joints. 

We appropriately modify the state of the art approaches to enable training using the introduced dataset. Our results demonstrate the effectiveness of the proposed dataset in enabling the modified approaches to overcome the aforementioned bottleneck and improve their performance, overall and on previously worst performing action categories. We also perform experiments to evaluate the relative importance of the introduced joints. We believe our contribution of new, expanded joint dataset will meet the twin objectives of improving performance and encouraging novel approaches in future. Going forward, we expect the research community to devise novel and efficient approaches for tackling dense skeleton representations present in our dataset. 

Our $118$ joints dataset, consisting of full body, fingers and even face joints can improve the recognition of actions based on expressions. This can help in capturing subtle changes in expression that would help in recognizing fine-grained actions (e.g. `moving head up' and `shaking head'). The significance of our work also arises from the emerging trend of fusing skeletal representations with other modalities (depth, RGB) for better performance in out of context, in-the-wild action recognition scenarios~\cite{mimetics, Gupta2021, moon2021integralaction}. The pretrained deep networks we introduce serve as a good starting point for such fusion based approaches.

\section{Acknowledgements}

(Portions of) the research in this paper used the NTU RGB+D (or NTU RGB+D 120) Action Recognition Dataset made available by the ROSE Lab at the Nanyang Technological University, Singapore. This material is based upon work supported by the Google Cloud Research Credits program with the award GCP19980904.

\bibliographystyle{ACM-Reference-Format}
\bibliography{ICVGIP21-CameraReady-Template}

% \appendix


% The appendix gets added after the references.


\end{document}